\documentclass[conference]{IEEEtran}
\IEEEoverridecommandlockouts
\usepackage{cite}
\usepackage{amsmath,amssymb,amsfonts}
\usepackage{algorithmic}
\usepackage{algorithm}
\usepackage{graphicx}
\usepackage{float} 
\usepackage{subfigure}
\usepackage{textcomp}
\usepackage{xcolor}
\usepackage{CJKutf8}
\usepackage[T1]{fontenc}
\usepackage{booktabs} % 用于创建高质量的 "三线表"
\usepackage{caption}  % 用于更好地控制标题样式
\usepackage{multirow}   % 用于处理跨行单元格
\newtheorem{lemma}{\textbf{Lemma}}
\def\BibTeX{{\rm B\kern-.05em{\sc i\kern-.025em b}\kern-.08em
		T\kern-.1667em\lower.7ex\hbox{E}\kern-.125emX}}
\begin{document}
\begin{CJK}{UTF8}{gbsn}

	\title{Multi-Task Learning with Feature-Similarity Laplacian Graphs for Predicting Alzheimer's Disease Progression
	\thanks{Corresponding author: Menghui Zhou, Po Yang}
	}
	
	\author{
		\IEEEauthorblockN{Zixiang Xu}
		\IEEEauthorblockA{\textit{School of Software} \\
			\textit{Yunnan University}\\
			Kunming, China \\
			zixiang.xu@foxmail.com}
		\and
		\IEEEauthorblockN{Menghui Zhou*}
		\IEEEauthorblockA{\textit{Department of Computer Science} \\
			\textit{The University of Sheffield}\\
			Sheffield, UK \\
			menghui.zhou@sheffield.ac.uk}
		\and
		\IEEEauthorblockN{Jun Qi}
		\IEEEauthorblockA{\textit{Department of Computing} \\
			\textit{\quad Xi’an JiaoTong-Liverpool University\quad}\\
			Suzhou, China \\
			Jun.Qi@xjtlu.edu.cn}
		\and
		\IEEEauthorblockN{Xuanhan Fan}
		\IEEEauthorblockA{\textit{School of Medical Technology} \\
			\textit{Beijing Institute of Technology}\\
			Beijing, China \\
			fanxuanhan@gmail.com}
		\and
		\IEEEauthorblockN{Yun Yang}
		\IEEEauthorblockA{\textit{School of Software} \\
			\textit{Yunnan University}\\
			Kunming, China \\
			yangyun@ynu.edu.cn}
		\and
		\IEEEauthorblockN{Po Yang*}
		\IEEEauthorblockA{\textit{Department of Computer Science} \\
			\textit{The University of Sheffield}\\
			Sheffield, UK \\
			po.yang@sheffield.ac.uk}
	}
	
	\maketitle
	
	\begin{abstract}
		Alzheimer's Disease (AD) is the most prevalent neurodegenerative disorder in aging populations, posing a significant and escalating burden on global healthcare systems. While Multi-Tusk Learning (MTL) has emerged as a powerful computational paradigm for modeling longitudinal AD data, existing frameworks do not account for the time-varying nature of feature correlations. To address this limitation, we propose a novel MTL framework, named Feature Similarity Laplacian graph Multi-Task Learning (MTL-FSL). Our framework introduces a novel Feature Similarity Laplacian (FSL) penalty that explicitly models the time-varying relationships between features. By simultaneously considering temporal smoothness among tasks and the dynamic correlations among features, our model enhances both predictive accuracy and biological interpretability. To solve the non-smooth optimization problem arising from our proposed penalty terms, we adopt the Alternating Direction Method of Multipliers (ADMM) algorithm. Experiments conducted on the Alzheimer’s Disease Neuroimaging Initiative (ADNI) dataset demonstrate that our proposed MTL-FSL framework achieves state-of-the-art performance, outperforming various baseline methods. The implementation source can be found at https://github.com/huatxxx/MTL-FSL.
	\end{abstract}
	
	% Furthermore, through longitudinal stability analysis, our model successfully identifies stable and informative biomarkers, highlighting its high interpretability and potential clinical value.
	
	
	\begin{figure*}[htbp]
		% 使用 \centering 命令来确保图片在整个页面宽度上居中
		\centering
		
		% \includegraphics 命令是插入图片的核心
		% width=\textwidth 表示图片的宽度将等于整个页面的文本宽度
		\includegraphics[width=\textwidth]{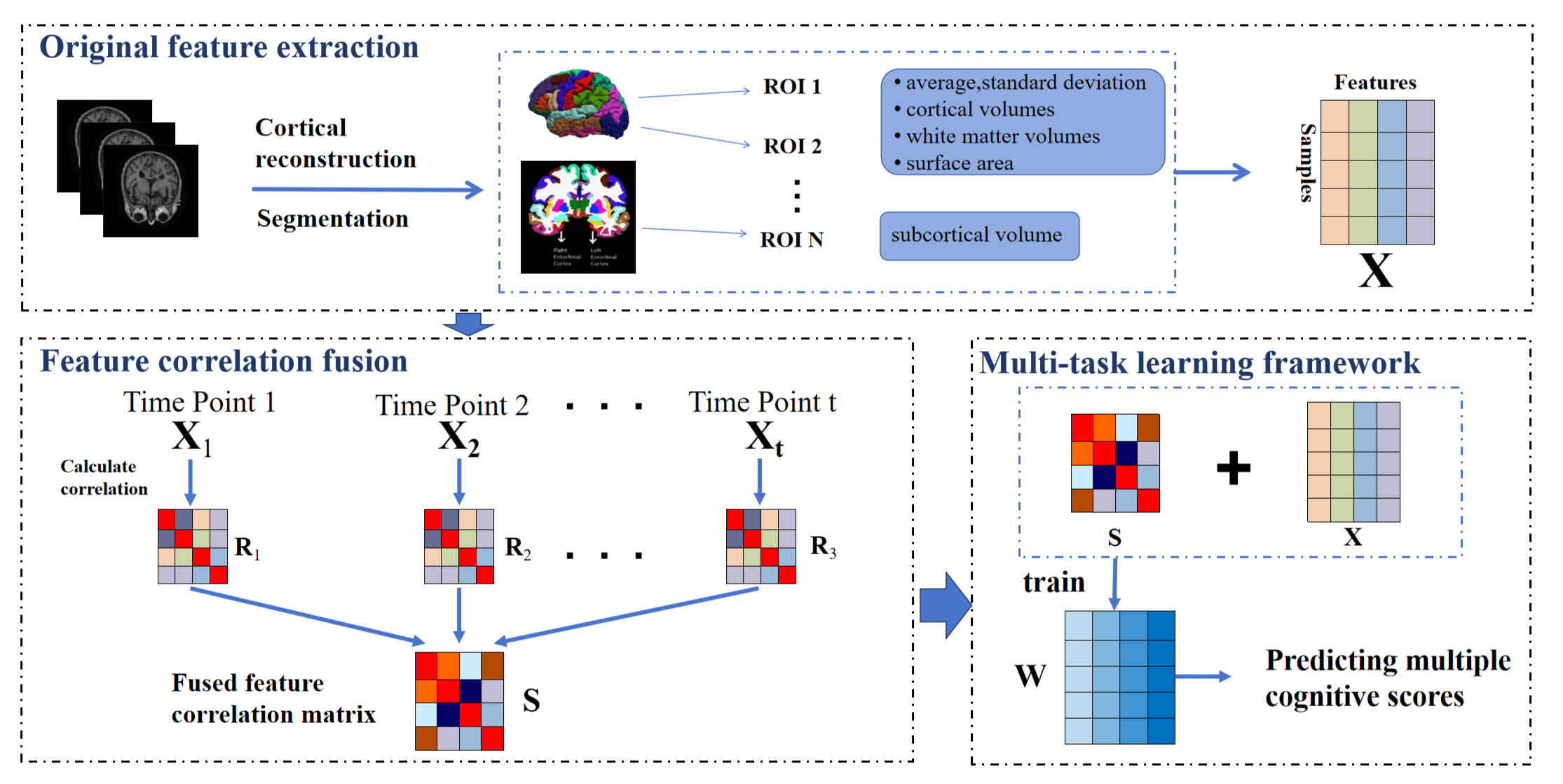} 
		
		% \caption 命令为图片添加标题
		\caption{The overall architecture of the Feature Similarity Laplacian graph Multi-Task Learning framework. The framework consists of two main components: a Feature Similarity Learning module that constructs a feature similarity graph, and a Multi-Task Learning module that leverages this graph to improve prediction performance.}
		
		% \label 命令为图片设置一个唯一的引用标签，必须放在 \caption 之后
		\label{fig:flowchart}
	\end{figure*}
	
	\section{Introduction}
	Alzheimer's disease (AD)\cite{b1} is the most common neurodegenerative disorder in aging societies and the leading cause of dementia. According to the World Health Organization (WHO)\cite{b2}, the number of people with dementia worldwide is projected to increase from 55 million in 2019 to 139 million by 2050 as the global population ages. Consequently, the annual costs associated with dementia care are expected to more than double, rising from \$1.3 trillion in 2019 to \$2.8 trillion by 2030. This poses a significant economic burden on healthcare systems and economies globally. Currently, there are no effective treatments to reverse or cure AD, and a definitive diagnosis relies on invasive procedures such as brain biopsy or autopsy. However, the pathological processes of AD begin years before the onset of overt clinical symptoms. Therefore, early diagnosis and accurate prediction of disease progression are crucial for designing clinical trials, developing intervention strategies, and formulating treatment plans. As such, accurately predicting disease progression and identifying the key biomarkers associated with it have become core objectives in AD research.  Previous studies have utilized various cognitive scores, such as the Mini-Mental State Examination (MMSE)\cite{MMSE} and the Alzheimer's Disease Assessment Scale-Cognitive sub-scale (ADAS-Cog)\cite{ADAS}, to assess the status of AD patients. These tools can be combined with MRI biomarkers to facilitate early diagnosis and predict the course of the disease\cite{b6}\cite{b7}. 
	
	In early research on predicting AD progression, traditional Single-Task Learning (STL) methods were predominantly employed. These approaches typically decompose the complex prediction problem into a series of independent sub-problems. However, the STL paradigm has inherent limitations.  Its core issue is that it treats the prediction at each time point as an isolated task, thereby ignoring the fundamental nature of AD as a neurodegenerative disease. The states of the disease at different time points are not mutually independent but are closely interconnected through intrinsic temporal correlations. STL methods fail to capture these evolving relationships in longitudinal data, leading to a waste of valuable information. This is particularly problematic in longitudinal AD studies where the number of subjects often decreases significantly over extended follow-up periods due to various reasons, such as withdrawal from the study. For future time points with sparse sample sizes, single-task models trained independently are highly prone to overfitting, which severely compromises the model's predictive performance and stability.
		
	To overcome the aforementioned limitations of single-task learning, the Multi-Task Learning (MTL)\cite{A survey on multi-task learning} paradigm has been introduced to the field of AD progression prediction and has proven to be a superior strategy. The core idea of MTL is to treat the prediction of cognitive scores at different time points as a series of related tasks and to learn all tasks within a unified framework. By incorporating regularization terms into the model's optimization objective that capture shared information across tasks, MTL models enable parameter sharing among different tasks during the learning process. This MTL paradigm is exceptionally well-suited for the demands of longitudinal AD research. By simultaneously leveraging data from all time points to train the model, MTL can effectively integrate information across different tasks, thereby significantly enhancing its generalization ability and predictive accuracy. This is particularly beneficial for distant time points where data is sparse; the model can achieve more effective parameter estimation by sharing information from data-richer, earlier time points, leading to more reliable predictions.
	
	Despite the significant success of Multi-Task Learning (MTL), existing methods still face key limitations in how they define and leverage "task relatedness. " A prominent issue is the adoption of overly strict or simplistic assumptions about the sharing mechanism for features. The majority of current MTL models, such as Temporal Group Lasso (TGL)\cite{TGL}, convex fused sparse group Lasso (cFSGL)\cite{cFSGL}, and MTL-LSA\cite{LSA}, typically employ an $\ell_{2,1}$-norm regularization. This term forces the prediction models for all time points  to share an identical subset of features. Under this "all-or-nothing" constraint, a biomarker is either considered contributory to the predictions at all time points or is deemed entirely irrelevant to all of them. While this strict assumption captures some of the commonality across tasks, it overlooks the dynamic nature of AD pathology. Biological research indicates that the importance of different types of biomarkers changes dynamically across the different stages of AD. Models like TGL are incapable of capturing these dynamic patterns, where a key feature might vary in importance over time, thus losing valuable temporal feature information. More importantly, this "hard" feature-sharing mechanism ignores the more complex correlations that exist among the biomarkers themselves. As a result, existing models lack explicit modeling of these inter-feature correlations.
	
	Therefore, how to construct a framework that can not only leverage the advantages of multi-task learning but also avoid overly simplistic or strict feature-sharing assumptions, thereby integrating the intrinsic correlations between features into the regularization term, is a critical step for enhancing the performance and biological interpretability of AD progression prediction models. Recognizing the importance of feature relationships in multi-task learning, we propose a novel MTL model that can simultaneously consider both the time-varying feature correlations and the task relatedness. Our model uses the  Feature Similarity Laplacian graph Multi-Task Learning framework shown in Fig.~\ref{fig:flowchart}, which fuses temporal smoothness relationships and multi-task feature correlations to improve predictive accuracy. This method constructs the associations between features into a graph structure and employs Laplacian regularization to maintain similarity between highly correlated features, thereby improving the predictive accuracy of our cognitive scores and the interpretability of the longitudinal biomarker selection.
	
	We summarize our contributions as follows:
	\begin{itemize}
	\item \textbf{Building a new MTL framework:} Our proposed MTL-FSL model is capable of capturing inter-feature correlations, which were not considered in previous models\cite{TGL}\cite{cFSGL}\cite{AutoTG}\cite{AMCOT}, and combines this with temporal smoothness. This enables our model to predict cognitive scores more accurately than various baseline methods.
	
	\item \textbf{Efficient optimization algorithm:} Considering the non-smoothness of the objective function introduced by our proposed regularization term, we design an optimization algorithm based on the Alternating Direction Method of Multipliers (ADMM). The advantage of ADMM lies in its ability to decompose the original problem into multiple sub-problems that are easier to solve. This allows for an efficient solution to the non-smooth convex optimization problem arising from the regularization, while ensuring convergence and improving computational performance.
	
	\item \textbf{Interpretable and Effective Methodology:} In contrast to current mainstream deep learning models, our method offers a significant advantage in terms of interpretability. Through longitudinal stability analysis, it can clearly reveal which MRI biomarkers exhibit persistent stability throughout the longitudinal progression of the disease. This provides significant clinical value and a scientific basis for biomarker discovery in Alzheimer's Disease.
	
	\end{itemize}
	
	\section{Related Work}
	A series of representative works have emerged in the domain of Multi-Task Learning (MTL) for longitudinal disease analysis.
	
	Regarding Deep Neural Networks (DNNs), particularly Recurrent Neural Networks (RNNs)\cite{RNNs} and Long Short-Term Memory (LSTM) networks, they have demonstrated powerful capabilities in processing time-series data, as they can effectively capture long-term dependencies. LSTMs, through their unique gating mechanisms, are capable of handling complex multivariate patterns. However, deep learning approaches also face inherent challenges. For instance, when addressing the common problem of missing values in clinical data, traditional imputation techniques can severely compromise model performance. Although research has begun to explore solutions, such as the forward-filling method combined with MinimalRNN proposed by Nguyen~et~al.~\cite{MinimalRNN}, and the end-to-end deep MTL framework designed by Liang~W.~et~al.~\cite{endto-end deep MTL} for adaptive imputation, the inherent uninterpretability of deep neural networks remains a fundamental limitation.
	
	In contrast, while traditional MTL methods may not be as proficient as deep models in capturing non-linear relationships, they exhibit a significant advantage in terms of interpretability. Zhou~et~al.~\cite{TGL} pioneered the temporal group lasso (TGL) model, which captures task relatedness by compelling all time points to share a common set of features. However, its drawback lies in neglecting the variability of biomarkers across different disease stages. To address this limitation, the subsequent cFSGL model~\cite{cFSGL} was developed as an improvement. It employs a sparse group lasso penalty to enable task-specific feature selection while maintaining temporal smoothness, thereby better reflecting the evolution of cognitive scores over time. In a further exploration of longitudinal analysis, M.~Zhou~et~al.~\cite{LSA} proposed a novel penalty termed Longitudinal Stability Adjustment (LSA) to adaptively capture the intrinsic, global temporal correlations across multi
	
	Yan~et~al.~\cite{G-SMuRFS} proposed a group $\ell_{2,1}$-norm to group related features from the same brain region, leveraging this prior knowledge for model training. The proposed model was termed Group-sparse Multi-task Regression and Feature Selection (G-SMuRFS). This approach, however, imposes a strict requirement on feature grouping, and such predefined groups may hinder the capture of implicit feature relationships. To address this issue, Tang~et~al.~\cite{FAS} introduced a multi-task model that simultaneously incorporates both task and feature relationship structures. By exploiting the implicit feature relationships among features from different brain regions at baseline, their model integrates inter-regional feature correlations. Nevertheless, these methods often model the system using only the feature correlations from the baseline time point, neglecting how these feature correlations evolve over time. Consequently, they also possess certain limitations.
	
	To address these limitations, we propose a novel feature penalty term, termed the  Feature Similarity Laplacian graph (FSL) penalty, which leverages the temporally evolving feature correlations to extract useful information among brain features. Building upon this, we introduce a multi-task learning framework, named MTL-FSL, that can simultaneously account for both the time-varying feature correlations and the task relationships. Our proposed framework achieves state-of-the-art prediction performance and is capable of identifying stable and highly informative biomarkers, thereby demonstrating a high degree of interpretability.
	\section{Methods}
	
	\subsection{Multi-task Learning}
	We formulate the AD cognitive score prediction problem within a Multi-Task Learning (MTL) framework, where the prediction of a cognitive score at each future time point is defined as a separate task. Consider a problem consisting of $t$ related tasks, where each task $i \in \{1, \dots, t\}$ is associated with a set of samples $(X_i, \mathit{y}_i)$. Our objective is to learn a parameter matrix $\mathit{W} = [\mathit{w}_1, \dots, \mathit{w}_t] \in \mathbb{R}^{p \times t}$, where each column vector $\mathit{w}_i \in \mathbb{R}^p$ represents the model parameters for the $i$-th task. Specifically, we assume all tasks share a common input data matrix $\mathit{X} \in [\mathit{X}_1, \dots, \mathit{X}_t] \in \mathbb{R}^{t \times (n \times p)}$, where each row of $\mathit{X}_i$ corresponds to a patient and each column corresponds to a feature. Correspondingly, the output matrix is denoted as $\mathit{Y} = [\mathit{y}_1, \dots, \mathit{y}_t] \in \mathbb{R}^{n \times t}$, where each column $\mathit{y}_i$ contains the cognitive scores of the $n$ patients at the $i$-th time point. We assume a linear relationship between the features and the targets and employ the squared loss function to quantify the prediction error. Therefore, the optimization objective of the model is to minimize the following loss function: $\mathit{L}(\mathit{Y}, \mathit{X}, \mathit{W}) = \frac{1}{2} \|\mathit{XW} - \mathit{Y}\|_F^2$.
	
	\subsection{Feature correlation}
	To comprehensively predict the progression of Alzheimer's Disease (AD), we can reasonably assume that two strongly correlated features should exhibit minimal differences in their model weights. We model this feature correlation using MRI features.
	First, to eliminate dimensional disparities and unify the data scale, we standardize the feature values at each time point into z-scores. Building upon this, to investigate the intrinsic relationships between features across different time points, we compute the Pearson correlation coefficient for each pair of features at every time point and construct the corresponding correlation matrices.
	
	\begin{equation}
		c_t^{m,l}=\frac{cov(X_t^m,X_t^l)}{\sigma_{X_t^m}\sigma_{X_t^l}}
	\end{equation}
	
	where $m, l \in \{1, \dots, p\}$. $X_{mt}$ denotes the value of the $m$-th feature at time point $t$. The larger $|c_{m,l}|$ indicates a stronger correlation between features $m$ and $l$. Based on these correlation coefficients, we construct a feature correlation matrix at each time point $t$, denoted as $\mathit{R}_t$.
	\begin{equation}
		R_t=
		\begin{bmatrix}
			c_t^{1,1} & c_t^{1,2} & \cdots & c_t^{1,p} \\
			c_t^{2,1} & c_t^{2,2} & \cdots & c_t^{2,p} \\
			\vdots & \vdots & \ddots & \vdots \\
			c_t^{p,1} & c_t^{p,2} & \cdots & c_t^{p,p}
		\end{bmatrix}
	\end{equation}
	
	\subsection{Feature Correlation Fusion Matrix}
	However, due to the varying number of patients at each time point, a simple averaging of these correlation matrices is not feasible. Therefore, we employ a weighted averaging approach to fuse the correlation matrices, where the weight for each time point is determined by the number of patients present at that time.
	We compute a weighted average of the feature correlations across multiple time points, and represent the associations between features as a graph structure. Then, using the method of constructing a Laplacian matrix, we create a fused feature correlation matrix $\mathit{S}$. By setting a threshold $\tau$, we stabilize the feature correlation matrix to focus only on highly correlated feature pairs. A connection is considered to exist between two features only when the absolute value of their correlation coefficient is greater than $\tau$. This method enables us to identify meaningful connections between different features. The specific construction process is shown in Algorithm~1.
	
	\begin{algorithm}[H]
		\caption{Constructing fused feature correlation matrix S}
		\renewcommand{\algorithmicrequire}{\textbf{Input:}}
		\renewcommand{\algorithmicensure}{\textbf{Output:}}
		
		\renewcommand{\algorithmicfor}{\textbf{for}}
		\renewcommand{\algorithmicdo}{\textbf{do}}
		\renewcommand{\algorithmicendfor}{\textbf{end for}}
		\begin{algorithmic}[1]
			\REQUIRE $X = \{X_1, X_2, \ldots, X_T\}, \tau$
			\ENSURE $S$
			\STATE Let $T$ be the number of time points in $X$.
			\STATE Let $d$ be the number of features.
			\STATE Initialize $R$ as an empty cell array and `patientcounts` as a zero vector.
			
			\FOR{$t = 1$ to $T$}
			\STATE $\text{patient\_counts}[t] \gets$ number of samples in $X_t$.
			\STATE $R_t \gets \text{corrcoef}(X_t)$.
			\STATE Set elements in $R_t$ where $|R_t| < \tau$ to 0.
			\ENDFOR
			
			\STATE Initialize $S$ as a $d \times d$ zero matrix.
			\FOR{$t = 1$ to $T$}
			\STATE $\text{weights[t]} \gets \text{patient\_counts[t]} / \sum(\text{patient\_counts})$.
			\STATE $S \gets S + \text{weights}[t] \times R_t$.
			\ENDFOR
		\end{algorithmic}
	\end{algorithm}
	
	\subsection{Feature Similarity Laplacian graph Penalty}

	By using the fused feature correlation matrix, we can more effectively capture features that vary over time. This enhances both the predictive accuracy of cognitive scores and the interpretability of the longitudinal biomarker selection.
	Therefore, we propose the following Feature Similarity Laplacian graph Penalty (FSL-Penalty):
	
	\begin{equation}
		\|{SW}\|_1 = \sum^{p}_{m,l = 1} |s_{m,l}| \|\mathbf{w}^m - \mathrm{sign}(s_{m,l})\mathbf{w}^l\|_1
	\end{equation}
	
	\subsection{One Novel MTL Framework}
	We employ the $\ell_1$-norm to ensure sparsity in the model's coefficient matrix. In the process of disease progression, it is reasonable to assume that the difference in cognitive scores between two consecutive time points is relatively small; therefore, we use a fused lasso penalty to incorporate temporal smoothness. Simultaneously, we integrate Feature Similarity Laplacian graph Penalty to capture the time-varying relationships among features.
	
	The novel multi-task learning framework with FSL-Penalty (MTL-FSL) is defined as:
	\begin{equation}
		\label{eq:MTL_FSL}
		\min_{\mathbf{w}}\frac{1}{2}\|Y-XW\|_{F}^{2}+\lambda_{1}\|W\|_{1}+\lambda_{2}\|SW\|_{1}+\lambda_{3}\|WH\|_{1}
	\end{equation}
	
	where $\lambda_1, \lambda_2, \lambda_3, \tau$ are all fine-tuned parameters, and ${H} \in \mathbb{R}^{T \times (T-1)}$ is defined as follows: $H_{ij} = 1$ if $i = j$, $H_{ij} = -1$ if $i = j+1$, $H_{ij} = 0$ otherwise.
	
	\section{Optimization Algorithms}
	
	By introducing slack variables $Q = W$, $P = SW$ and $V=WH$, Eq. \ref{eq:MTL_FSL} can be rewritten in ADMM form as Eq. \ref{eq:ADMM_MTL_FSL}.
		
	\begin{equation}
		\label{eq:ADMM_MTL_FSL}
		\begin{aligned}
			&\min_{W,Q,P,V}\frac{1}{2}\|Y-XW\|_F^2+\lambda_1\|Q\|_1+\lambda_2\|P\|_1+\lambda_3\|V\|_1\\
			&\mathrm{s.t.}\quad W-Q=0,SW-P=0,WH-V=0.
		\end{aligned}
	\end{equation}
	
	The augmented Lagrangian of Eq. \ref{eq:Lagrangian_W} is:
	
	\begin{equation}
		\label{eq:Lagrangian_W}
		\begin{aligned}
			\mathcal{L}_{\rho}(W, Q, P, V, &U_q, U_p, U_v) =\frac{1}{2}\|Y - XW\|_{F}^{2} \\
			&+ \lambda_1\|Q\|_{1} + \lambda_2\|P\|_{1} + \lambda_3\|V\|_{1} \\
			&+ \langle U_q, W - Q \rangle + \frac{\rho}{2}\|W - Q\|^{2} \\
			&+ \langle U_p, SW - P \rangle + \frac{\rho}{2}\|SW - P\|^{2} \\
			&+ \langle U_v, WH - V \rangle + \frac{\rho}{2}\|WH - V\|^{2}.
		\end{aligned}
	\end{equation}
	
	\textbf{Update \textsl{W}}: The update of $W$ at the $(t+1)$-th iteration is obtained from Eq. \ref{eq:update_W}.
	
	\begin{equation}
		\label{eq:update_W}
		\begin{aligned}
			W^{(t+1)} &={} \arg \min_{W} \frac{1}{2} \|Y - XW\|_{F}^{2} \\
			& + \langle U_{q}^{(t)}, W - Q^{(t)} \rangle + \frac{\rho}{2} \|W - Q^{(t)}\|^{2} \\
			& + \langle U_{p}^{(t)}, SW - P^{(t)} \rangle + \frac{\rho}{2} \|SW - P^{(t)}\|^{2} \\
			& + \langle U_{v}^{(t)}, WH - V^{(t)} \rangle + \frac{\rho}{2} \|WH - V^{(t)}\|^{2} 
		\end{aligned}
	\end{equation}
	
	To find the closed-form solution for $W$, we take the partial derivative of Eq. \ref{eq:update_W} with respect to $W$ and set it to zero, which yields Eq. \ref{eq:dW}.
	
	\begin{equation}
		\label{eq:dW}
		\begin{aligned}
		0 = -X^{\mathrm{T}}(Y - XW) &+ U_{q}^{(t)} + \rho(W - Q^{(t)}) \\
		&+ S U_{p}^{(t)} + \rho S(SW - P^{(t)})\\
		&+ U_{v}^{(t)} H^\mathrm{T} + \rho (WH - V^{(t)}) H^\mathrm{T}.
		\end{aligned}
	\end{equation}
	
	We rewrite the equation into the following form, placing the terms related to $W$ on one side:
	\begin{equation}
		\begin{aligned}
		(X^{\mathrm{T}} X + \rho I &+ \rho S S)W + \rho W H H^{\mathrm{T}} = \\
		&X^{\mathrm{T}} Y - U_q^{(t)} + \rho Q^{(t)} - S U_p^{(t)} + \rho S P^{(t)} \\
		&- U_v^{(t)} H^{\mathrm{T}} + \rho V^{(t)} H^{\mathrm{T}}.
		\end{aligned}
	\end{equation}
	
	To simplify the solution, we first introduce the following auxiliary variables:
	
	\begin{align*}
		C^{(t)} &= X^{\mathrm{T}} Y - U_q^{(t)} + \rho Q^{(t)} - S U_p^{(t)} \\
		&\qquad+ \rho S P^{(t)} - U_v^{(t)} H^{\mathrm{T}} + \rho V H^{\mathrm{T}}, \\
		F &= HH^{\mathrm{T}}, \\
		L_i &= X_i^{\mathrm{T}} X_i + \rho I + \rho S S.
	\end{align*}
	
	To efficiently update $W$ using the Cholesky decomposition, we decompose $W$ column-wise. The above matrix equation can then be decoupled into $t$ independent vector equations, one for each column $i=1, 2, \dots, t$:
	\begin{equation}
		\begin{aligned}
			L_i w_i = c_i - \rho W f_i.
		\end{aligned}
	\end{equation}
	Clearly, $L_i, i \in {1, \dots, t} $ is symmetric positive definite, which Cholesky factorization is applicable for, resulting in efficient updating of $W$ .
	
	\textbf{Update} \textbf{\textit{Q}}, \textbf{\textit{P}}, \textbf{\textit{V}}:Updating $Q$ needs to solve the following problems
	
	\begin{equation}
		\begin{aligned}
			Q^{(t+1)}=\arg\min_Q\lambda_1&\|Q\|_1+<U_q^{(t)},W^{(t+1)}-Q>\\
			&+\frac{\rho}{2}\|W^{(t+1)}-Q\|^2,
		\end{aligned}
	\end{equation}
	
	which is equivalent to the following problem:
	
	\begin{equation}
		\label{eq:update_Q}
		Q^{(t+1)}=\arg\min_Q\frac{1}{2}\|Q- \Theta^{(t+1)}\|^2+\frac{\lambda_1}{\rho}\|Q\|_1
	\end{equation}
	where $\Theta^{(t+1)} = \left( W^{(t+1)} + U_q^{(t)}\right)$. Then we can use the following lemma to update $Q^{(t+1)}$ \cite{FIST} .
	
	\begin{lemma}
		For any $\lambda_1 \geq 0$, we can calculate Eq.\eqref{eq:update_Q} by the following:
		\begin{equation} \label{eq:24}
			q_{ij}^{(t+1)} = \operatorname{sign}\left(\theta_{ij}^{(t+1)}\right) \max\left(\left|\theta_{ij}^{(t+1)}\right| - \frac{\lambda_1}{\rho}, 0\right),
		\end{equation}
		where $q_{ij}^{(t+1)}$ and $\theta_{ij}^{(t+1)}$ denote the element of matrix $Q^{(t+1)}$ and $\Theta^{(t+1)}$ respectively.
	\end{lemma}
	
	Updating $P$, $V$ needs to solve the following problems
	
	\begin{equation}
		\label{eq:update_P}
		P^{(t+1)}=\arg\min_P\frac{1}{2}\|P-  \left( S W^{(t+1)} + U_p^{(t)}\right)\|^2+\frac{\lambda_2}{\rho}\|P\|_1
	\end{equation}
	
	\begin{equation}
		\label{eq:update_V}
		V^{(t+1)}=\arg\min_V\frac{1}{2}\|V- \left( W^{(t+1)}H + U_v^{(t)}\right)\|^2+\frac{\lambda_3 	}{\rho}\|V\|_1
	\end{equation}
	
	The updates for variables $P$ and $V$ are similar to that of $Q$; therefore, the details are omitted for brevity.
	
	\textbf{Update} \textbf{\textit{U}${_\textit{q}}$}, \textbf{\textit{U}$_\textit{p}$}, \textbf{\textit{U}$_\textit{v}$}: According to the standard ADMM, the updates of augmented lagrangian multipliers are as follows:
	
	\begin{equation}
		\label{eq:Update_U}
		\begin{aligned}
			U_q^{(t+1)}&=\quad U_q^{(t)}+\rho\left(W^{(t+1)}-Q^{(t+1)}\right),\\
			U_p^{(t+1)}&=\quad U_p^{(t)}+\rho\left(SW^{(t+1)}-P^{(t+1)}\right),\\
			U_v^{(t+1)}&=\quad U_v^{(t)}+\rho\left(W^{(t+1)}H-V^{(t+1)}\right).
		\end{aligned}
	\end{equation}
	
	Taken together, the entire solution procedure based on the ADMM algorithm is presented in Algorithm 2.
	
	\begin{algorithm}[H]
		\caption{ADMM optimization of FAS-MTFL}
		\label{alg:admm}
		\textbf{Input:} $X$, $Y$, $\lambda_1$, $\lambda_2$, $\lambda_3$ $\rho$. \\
		\textbf{Output:} $W$.
		\renewcommand{\algorithmicfor}{\textbf{for}}
		\renewcommand{\algorithmicdo}{\textbf{do}}
		\renewcommand{\algorithmicendfor}{\textbf{end for}}
		\begin{algorithmic}[1]
			\STATE \textbf{Initialization:} $W^{(0)} \leftarrow 0$, $Q^{(0)} \leftarrow 0$, $P^{(0)} \leftarrow 0$, $V^{(0)} \leftarrow 0$, $U_{q}^{(0)} \leftarrow 0$, $U_{p}^{(0)} \leftarrow 0$, $U_{v}^{(0)} \leftarrow 0$.
			\REPEAT
			\STATE Update $W^{(t+1)}$ according to Eq.~\eqref{eq:update_W}.
			\STATE Update $Q^{(t+1)}$ according to Eq.~\eqref{eq:update_Q}.
			\STATE Update $P^{(t+1)}$ according to Eq.~\eqref{eq:update_P}.
			\STATE Update $V^{(t+1)}$ according to Eq.~\eqref{eq:update_V}.
			\STATE Update $U_{q}^{(t+1)}$, $U_{p}^{(t+1)}$, $U_{v}^{(t+1)}$ according to Eq.~\eqref{eq:Update_U}.
			\UNTIL{ Stopping criterion is satisfied..}
		\end{algorithmic}
	\end{algorithm}
	
\begin{figure}[htbp]
		% 使用 \centering 命令来确保图片在整个页面宽度上居中
		\centering

		\includegraphics[scale=0.4]{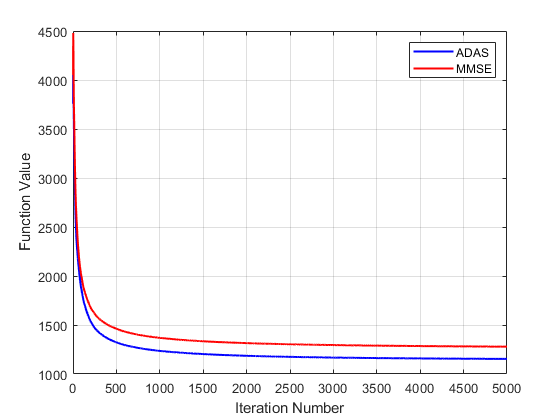} 
		
		% \caption 命令为图片添加标题
		\caption{The convergence situation of using  ADMM algorithm in the ADAS and MMSE datasets.}
		
		% \label 命令为图片设置一个唯一的引用标签，必须放在 \caption 之后
		\label{fig:admm_convergence}
	\end{figure}

	Fig.~\ref{fig:admm_convergence} illustrates the convergence of the ADMM algorithm on the ADAS and MMSE datasets, plotting the objective function value against the iteration number. The curves for both tasks show a rapid initial decrease, followed by a gradual stabilization. The algorithm effectively converges after approximately 1500 iterations, which validates the stability and efficiency of our chosen optimization strategy.
	
	\section{Experimental Result}
	
	The data used in this study were obtained from the publicly available Alzheimer’s Disease Neuroimaging Initiative (ADNI) database. The ADNI dataset comprises longitudinal observations spanning up to 120 months, with data collected on a semi-annual or annual basis. To focus on the early stages of the disease, we utilized the data from the first 48 months and applied the following preprocessing steps: First, we excluded subjects who lacked baseline MRI records. Second, MRI features that failed quality control (QC) were removed. Finally, mean imputation was employed to fill the missing values in the remaining features. 
	
	Following these steps, our final analytical dataset consisted of 314 features across 6 time points, as summarized in Table~\ref{tab:adni_summary}.
	
	\begin{table}[htbp]
		\centering
		\caption{Demographic Information of Subjects at Different Time Points}
		\label{tab:adni_summary}
		\begin{tabular*}{\columnwidth}{@{\extracolsep{\fill}}lcccc@{}}
			\toprule
			\textbf{Time point} & \multicolumn{3}{c}{\textbf{Category}} & \textbf{Total} \\
			\cmidrule(lr){2-4}
			& AD   & MCI   &  CN  &       \\
			\midrule
			\textbf{Baseline}   & 315  & 817   & 400  & 1532   \\
			\textbf{Month 6}    & 265  & 669   & 383  & 1317   \\
			\textbf{Month 12}   & 366  & 768   & 241  & 1375   \\
			\textbf{Month 24}   & 132  & 653   & 314  & 1099   \\
			\textbf{Month 36}   & 0    & 277   & 155  & 432   \\
			\textbf{Month 48}   & 0    & 172   & 117  & 432   \\
			\bottomrule
		\end{tabular*}
	\end{table}
	
	\subsection{Model Training and Evaluation} % Or use \subsubsection if appropriate
	
	For the model training and evaluation, we partitioned the dataset into a training set (90\%) and a testing set (10\%). A 10-fold cross-validation strategy was employed on the training set to select the optimal regularization parameters. The performance of the model was assessed using the following three metrics:	
	
	\begin{itemize}
		\item \textbf{Root Mean Squared Error (rMSE):} 
		\begin{equation*}\mathrm{rMSE}(y,\hat{y})=\sqrt{\frac{\|y-\hat{y}\|_2^2}{n}},\end{equation*}
		\item \textbf{Normalized Mean Squared Error (nMSE):} 
		\begin{equation*}\mathrm{nMSE}\left(Y,\hat{Y}\right)=\frac{\sum_{i=1}^t\left\|Y_i-,\hat{Y}_i\right\|_2^2/\sigma(Y_i)}{\sum_{i=1}^tn_i}\end{equation*}
		\item \textbf{Weighted Correlation Coefficient (wR):} 
		\begin{equation*}\mathrm{wR}\left(Y,\hat{Y}\right)=\frac{\sum_{i=1}^t\mathrm{Corr}\left(Y_i,\hat{Y}_i\right)n_i}{\sum_{i=1}^tn_i},\end{equation*}
	\end{itemize}
	
	Our objective was to achieve lower RMSE and nMSE values and a higher wR value, which would indicate the superiority of our proposed model.

	\subsection{Comparison of Models}
	
	We evaluated our proposed algorithm on the ADNI dataset, with all experiments implemented in MATLAB. The performance of our method was benchmarked against several state-of-the-art models, including: TGL~\cite{TGL}, cFSGL~\cite{cFSGL}, AutoTG~\cite{AutoTG}, LSA~\cite{LSA}, AMCOT~\cite{AMCOT}, FL-SGL~\cite{FL-SGL}, MinimalRNN~\cite{MinimalRNN}, and FAS~\cite{FAS}.	The regularization parameters for all baseline methods were tuned over the range $\lambda \in \{0.01, 0.1, 1, 10, 50, 100, 500, 1000\}$. For our proposed FSL-Penalty, the corresponding parameter $\lambda$ was selected from the set $\{0.01, 0.02, \dots, 0.1\}$.
	
	The experimental results for predicting ADAS and MMSE scores are summarized in Table~\ref{tab:comparison_results}. The table clearly demonstrates the superior performance of our proposed model, FSL(Ours), across all evaluated metrics.
	
	Specifically, for the ADAS score prediction, our FSL model achieves an nMSE of $\mathbf{0.468 \pm 0.023}$ and a wR of $\mathbf{0.734 \pm 0.017}$. These results are notably better than the second-best performing methods, FL-SGL (nMSE of $0.469 \pm 0.018$) and cFSGL (wR of $0.728 \pm 0.021$). A similar trend is observed for the MMSE score prediction, where our model again secures the best performance with an nMSE of $\mathbf{0.520 \pm 0.030}$ and a wR of $\mathbf{0.695 \pm 0.0214}$.
	
	It is worth noting that our method consistently outperforms a diverse range of baseline models, including traditional MTL approaches (e.g., TGL, cFSGL, LSA) and recent deep learning-based methods (e.g., MinimalRNN). This comprehensive superiority underscores the effectiveness of our proposed feature similarity learning penalty in capturing meaningful relationships for longitudinal data analysis.
	
	\begin{table}[htbp]
		\centering
		% --- 可选：定制标题以匹配图片样式 ---
		\caption{Demographic Information of Subjects at Different Time Points}
		\label{tab:comparison_results}
		
		% --- 关键：增加行高 ---
		\renewcommand{\arraystretch}{1.3} % 调整这个数值来改变行间距，1.0 是默认值
		
		\resizebox{1.0\linewidth}{!}{
			\begin{tabular}{lcccc} 
				\toprule
				\multirow{2}{*}{\textbf{Method}} & \multicolumn{2}{c}{\textbf{ADAS}} & \multicolumn{2}{c}{\textbf{MMSE}} \\
				\cmidrule(lr){2-3} \cmidrule(lr){4-5}
				& nMSE   & wR   & nMSE   & wR     \\
				\midrule
				\textbf{TGL}   & $0.495 \pm 0.030$  & $0.714 \pm 0.020$   & $0.556 \pm 0.031$  & $0.667 \pm 0.025$   \\
				\textbf{cFSGL}    & $0.474 \pm 0.030$  & $0.728 \pm 0.021$   & $0.529 \pm 0.026$  & $0.690 \pm 0.017 $  \\
				\textbf{AutoTG}   & $0.500 \pm 0.033$  & $0.714 \pm 0.023$   & $0.543 \pm 0.032$  & $0.683 \pm 0.020$   \\
				\textbf{LSA}   & $0.506 \pm 0.036$  & $0.705 \pm 0.027$   & $0.596 \pm 0.030$  & $0.642 \pm 0.025$   \\
				\textbf{AMCOT}   & $0.484 \pm 0.028$  & $0.724 \pm 0.018$   & $0.542 \pm 0.019$  & $0.681 \pm 0.015$   \\
				\textbf{FL-SGL}   & $0.469 \pm 0.018$  & $0.733 \pm 0.013$   & $0.529 \pm 0.032$  & $0.691 \pm 0.022$   \\
				\textbf{FAS}   & $0.492 \pm 0.029$  & $0.714 \pm 0.020$   & $0.590 \pm 0.030$  & $0.643 \pm 0.024$   \\
				\textbf{MinimalRNN}   & $0.512 \pm 0.132$  & $0.690 \pm 0.093$   & $0.692 \pm 0.115$  & $0.532 \pm 0.127$   \\
				\textbf{FSL(Ours)}   & $\mathbf{0.468 \pm 0.023}$  & $\mathbf{0.734 \pm 0.017}$   & $\mathbf{0.520 \pm 0.030}$  & $\mathbf{0.695 \pm 0.0214}$  \\
				\bottomrule
			\end{tabular}
		}
		
	\end{table}
	
	\subsection{Longitudinal Prediction Performance}
	
\begin{figure}[H]
		\centering  %图片全局居中
		\subfigure[ADAS]{
			\label{Fig.ADAS}
			\includegraphics[width=0.48\columnwidth]{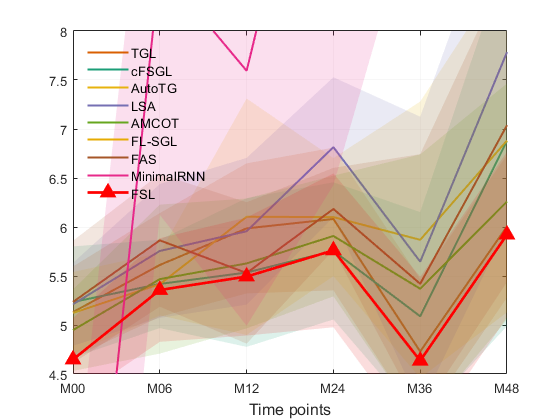}}
		\subfigure[MMSE]{
			\label{Fig.MMSE}
			\includegraphics[width=0.48\columnwidth]{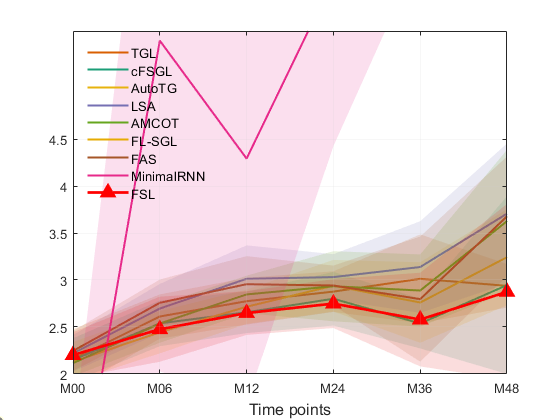}}
		\caption{The average root Mean Squared Error (rMSE) compared to the baseline model is presented.}
		\label{Fig.main}
	\end{figure}

	Figures~\ref{Fig.ADAS} and \ref{Fig.MMSE} illustrate the longitudinal prediction performance of our proposed FSL model against several baseline methods on the ADAS and MMSE cognitive scores, respectively. The y-axis in these plots represents the rMSE, where lower values indicate better performance. A clear trend is observable across both figures: our proposed FSL model (highlighted by the solid red line with triangle markers) demonstrates a superior performance trend, achieving the lowest or near-lowest prediction error at most time points from baseline (M00) to 48 months (M48).
	
	In the ADAS prediction task (Fig.~\ref{Fig.ADAS}), the FSL model largely maintains the lowest error trajectory and exhibits remarkable stability, especially at the crucial early and mid-stage time points (M06, M12, M24, M36). This is further evidenced by the narrower shaded area around the FSL curve, which represents the standard deviation of the results, indicating a lower variance and higher robustness compared to most competing methods.
	
	This overall superiority is also evident in the MMSE prediction task (Fig.~\ref{Fig.MMSE}). While other methods may perform comparably at isolated time points, our FSL model maintains a consistently low error profile throughout the entire follow-up period. Its advantage becomes particularly pronounced at later stages (M36 and M48), suggesting that our model is more effective at handling the increased difficulty and data sparsity associated with long-term prediction.

	\subsection{The Identification of Longitudinal MRI Biomarkers}
	
\begin{figure}[H]
		\centering  %图片全局居中
		\subfigure[Framework:MTL-FSL, Feature:MRI, Target:ADAS (24 stable features)]{
			\label{Fig.ADAS_MRI}
			\includegraphics[width=0.8\columnwidth]{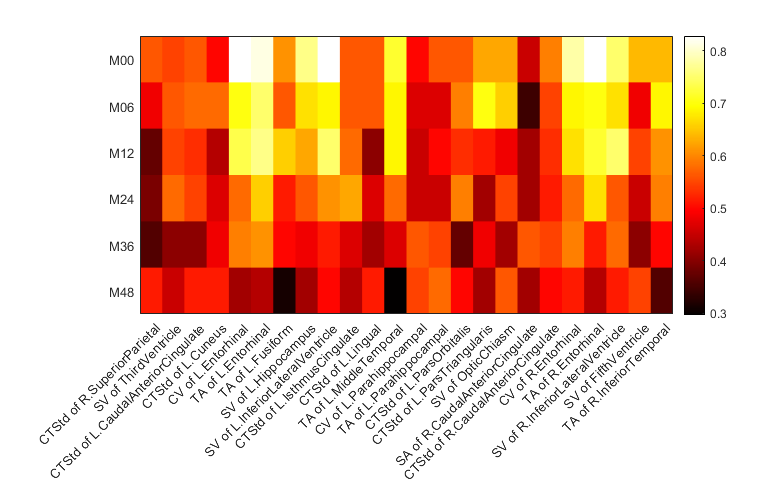}}
		\subfigure[Framework:MTL-FSL, Feature:MRI, Target:MMSE (29 stable features)]{
			\label{Fig.MMSE_MRI}
			\includegraphics[width=0.8\columnwidth]{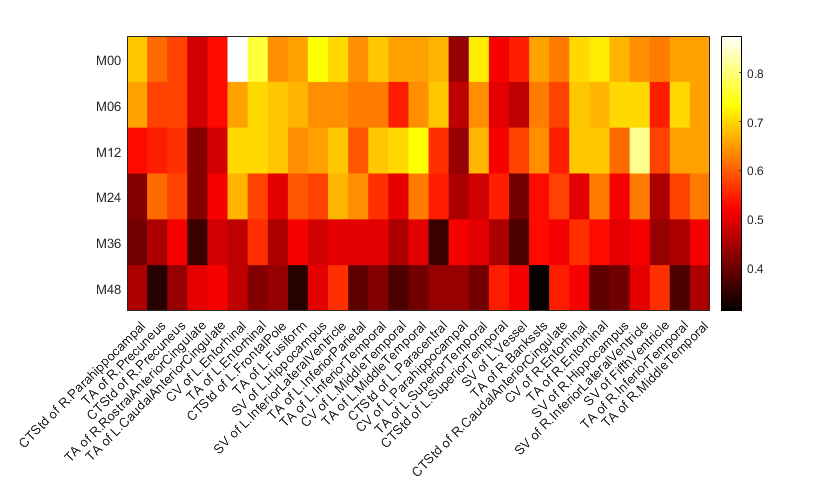}}
		\caption{The stability selection results of the ADAS and MMSE prediction tasks.}
		\label{Fig.heatmap}
	\end{figure}

	One of the key strengths of our MTL-FSL framework is its ability to identify temporally significant biomarkers. To investigate this, we employ the longitudinal stability selection method~\cite{cFSGL} to analyze the MRI Obiomarkers identified by our model. Our analysis covers the initial 48 months, where our model demonstrated robust performance. The stability selection results for the ADAS and MMSE prediction tasks are presented in Fig.~\ref{Fig.heatmap}. 
	
	The heatmaps highlight several brain regions known to be critical in AD pathology. As shown in Fig.~\ref{Fig.ADAS_MRI}, for the ADAS task, our model identifies 24 stable features. Notably, features associated with the left entorhinal cortex and the left hippocampusdemonstrate high stability, particularly in the early stages. The volume of the left hippocampus is a well-established biomarker in numerous AD studies~\cite{placeholder_hippocampus_ref}, with significant atrophy in this region linked to the memory deficits characteristic of the prodromal phase of AD. Our model also captures the sustained importance of ventricle enlargement, as evidenced by the high stability of the SV of ThirdVentricle and the SV of L.InferiorLateralVentricle. These ventricle changes often serve as sensitive indicators for assessing the progression from MCI to AD~\cite{placeholder_ventricle_ref}. Fig.~\ref{Fig.MMSE_MRI} presents the 29 stable biomarkers identified for the MMSE task. While there is an overlap of key regions like the bilateral entorhinal cortex and the hippocampus, distinct patterns also emerge. For instance, the thickness of the right middle temporal gyrus and the thickness of the left fusiform gyrus are identified as highly stable biomarkers for MMSE, aligning with literature that associates the inferotemporal cortex with memory performance and cognitive function~\cite{placeholder_inferotemporal_ref}. They serve as vital biomarkers for tracking disease progression and have the potential to aid in the evaluation of therapeutic responses.

	\section{Conclusion}
	
	In this paper, we addressed the prediction of AD progression by introducing a novel framework, MTL-FSL, which explicitly models the dynamic, time-varying correlations among features from longitudinal data. Our model, by incorporating a novel Feature Similarity Laplacian (FSL) penalty, demonstrated state-of-the-art performance, outperforming a wide range of baseline methods on the ADNI dataset across multiple cognitive targets. To handle the non-smooth, convex objective function introduced by our regularization terms, we developed an efficient optimization algorithm based on the Alternating Direction Method of Multipliers (ADMM), which decomposes the problem into several easier-to-solve sub-problems. For future work, we plan to extend the application of MTL-FSL to other neurodegenerative diseases, such as Parkinson's disease, to validate its generalizability. Furthermore, we will explore the integration of multi-modal data, such as genetic and PET imaging data, to further enhance the model's predictive power and interpretability.
	
	\section*{Acknowledgement}
	
	This research was supported by the National Natural Science Foundation of China (No.62301452).

	\vspace{12pt}
\end{CJK}
\end{document}